\documentclass[12pt,letterpaper]{amsart}
\usepackage{hyperref}
\usepackage{graphicx, color}
\usepackage{mathtools}
\usepackage{enumerate,enumitem}
\usepackage{bbm}
\usepackage[margin=1in]{geometry}
\usepackage{amssymb,amsthm}
\usepackage{subcaption}
\usepackage{float}
\usepackage{tikz}
\usepackage{yhmath}
\usepackage[numbers]{natbib}
\usepackage{tabu}
\usepackage{tabularx}
\usepackage{algorithm}
\usepackage{algorithmicx}
\usepackage{algpseudocode}
\usepackage{multirow}
\usepackage{makecell}

\usepackage{mathabx}

\usepackage[flushleft]{threeparttable}
\usepackage{array,booktabs,makecell}
\usepackage{microtype}
\usepackage{nicematrix}
\usepackage{arydshln}

\usepackage[capitalize,noabbrev]{cleveref}

\newtheorem{theorem}{Theorem}[section]
\newtheorem{proposition}[theorem]{Proposition}

\newtheorem{corollary}[theorem]{Corollary}
\theoremstyle{definition}

\theoremstyle{remark}

\DeclareMathOperator{\argmax}{argmax}

\newcommand{\R}{\mathbb{R}}

\newcommand{\A}{\mathcal{A}}

\definecolor{OliveGreen}{rgb}{0,0.6,0}

\title{Beacon: Post-Training Quantization with Integrated Grid Selection}

 \author{Shihao Zhang, Rayan Saab} 
 \thanks{This work was partially supported by NSF grant DMS-2410717.}%
\thanks{Shihao Zhang is with the Department of Mathematics, UC San Diego (shz051@ucsd.edu). Rayan Saab is with the Department of Mathematics and HDSI, UC San Diego (rsaab@ucsd.edu).}

\begin{document}
\maketitle
\begin{abstract}
    Quantization is a widely used compression technique for reducing the memory and computation costs of large pre-trained models. A key challenge in per-channel post-training quantization (PTQ) is selecting appropriate scaling factors to replace weight values with values from a scaled integer grid. Existing methods typically fix the scale at the outset via heuristic tuning or grid search. We propose {\bf Beacon}, a simple and effective algorithm that eliminates the need for such manual tuning. Beacon performs per-channel PTQ directly using an unscaled grid and automatically determines the optimal scaling factors by exploiting the geometry of scalar quantization. It  does not rely on back-propagation or large calibration sets. Despite its simplicity and tuning-free nature, Beacon achieves competitive performance compared to state-of-the-art methods, making it a practical solution for efficient model deployment.
\end{abstract}

\section{Introduction}
Recent deep neural networks—most notably large language models (LLMs)—have massive computational and memory demands. Quantization \cite{gholami2022survey, liang2021pruning, wei2024advances} has emerged as a mainstream compression technique \cite{8253600, deng2020model, zhang2025theoretical} for deploying LLMs on resource-constrained devices and in other resource-constrained settings. By reducing the number of bits, or  bit width, used to represent weights or activations, quantization lowers storage requirements, memory bandwidth, and computation costs.
Post-training quantization (PTQ) \cite{frantar2022gptq, zhang2023post, cheng2024optimize, zhang2025provable} is a particularly attractive approach for its simplicity. It avoids backpropagation and adapts pre-trained models usually in a single pass or in very few passes over a small calibration data set. So, it incurs far less computational overhead than quantization-aware training (QAT) \cite{Jacob_2018_CVPR, xi2023training, zhang2022learning}, where quantized models are trained directly with gradient-based methods.

In practice, PTQ is most commonly implemented with scalar quantization, where each weight is ultimately represented by a value from a finite scalar alphabet, such as a uniform grid. While modern scalar PTQ algorithms such as GPTQ \cite{frantar2022gptq}, GPFQ \cite{lybrand2021greedy, zhang2023post}, and Qronos \cite{zhang2025qronos}  incorporate dependencies across weights during the quantization process, the quantization function itself still acts coordinatewise, mapping real values to scalar representatives. Scalar methods have become the industry standard and perform reliably at bit widths of 4 or more. However, pushing them into the ultra-low bit regime ($<3$ bits) remains difficult, even with transformation-based enhancements \cite{xiao2023smoothquant, lin2024awq, ashkboos2024quarot, liu2024spinquant}. An alternative is vector quantization (VQ), where groups of weights are jointly mapped to codewords from a vector codebook rather than to a scalar alphabet. VQ offers accuracy gains at ultra-low precision but at the expense of higher complexity and reduced deployment efficiency.
In contrast, our method remains within the scalar quantization paradigm, with its minimal inference overhead, while achieving strong accuracy improvements at ultra-low bit widths.

{\bf Notation.} We denote the weight matrix of a layer by $W \in \mathbb{R}^{N \times N'}$, where each of the $N'$ columns is an $N$-dimensional channel $w \in \mathbb{R}^N$. Given a vector $v \in \mathbb{R}^n$, we use $v_i$ for its $i$-th entry, $v_{\geq j}$ for the subvector $(v_j, \dots, v_n)^\top$, and we define $v_{\leq j}$ analogously. $\|v\|$ is the Euclidean norm of $v$. Given a matrix $A \in \mathbb{R}^{m \times n}$, we use $A_i$ to denote its $i$-th column. We use $A_{\leq j}$ to denote the submatrix $(A_1, \dots, A_j)$.

The standard $b$-bit integer grid is $\{0,1,\dots,2^b\!-\!1\}$, and weight quantization uses its scaled and shifted version
\(
\mathcal{A}_c = \{ c \cdot (z+k) : k=0,\dots,2^b\!-\!1 \},
\)
where $c$ is the scaling factor and $z$ is the offset (zero point). In asymmetric quantization, $c$ is typically defined per channel on a scaled min–max grid,
\(
c = \frac{\alpha \cdot \max(w) - \beta \cdot \min(w)}{2^b-1}.
\)
The associated round-to-nearest (RTN) operator $\mathcal{Q}$ is
\begin{equation*}
\mathcal{Q}(\star) = c \cdot \left( \text{clip}\!\left( \left\lceil \tfrac{\star}{c} - z \right\rfloor ; 0, 2^b-1 \right) + z \right),
\end{equation*}
where $\text{clip}(x;a_{\min},a_{\max}) = \min\{\max\{x,a_{\min}\},a_{\max}\}$.

This paper considers asymmetric per-channel PTQ. For each channel (column) $w \in \mathbb{R}^N$ of $W$, we \emph{fix} its unscaled integer grid
\(
\mathcal{A} = \{z+k : k=0,\dots,2^b\!-\!1\},
\)
where we follow the standard choice of zero point \begin{equation}\label{eq:zero point}
    z=\left\lceil\frac{\min(w)}{\max(w)-\min(w)} \cdot (2^b-1)\right\rfloor
\end{equation}
throughout this paper. Each channel is quantized using a scaling factor $c \in \mathbb{R}$. Collecting these gives a vector $s \in \mathbb{R}^{N'}$ of per-channel scaling factors for $W$. 

{\bf Related Work.} In per-channel PTQ, accurate scaling is critical for preserving model quality, particularly at ultra-low bit widths. Most scalar PTQ rounding methods \cite{zhang2023post, frantar2022gptq, cheng2024optimize} determine $s$ by tuning $\alpha$ and $\beta$ once at initialization and keeping them fixed thereafter. A recent exception is the method of \citet{zhang2025comq}, which updates $s$ during its iterations but is highly sensitive to the initial choices of $\alpha$ and $\beta$. Other approaches attempt to refine scaling through trial-and-error, via grid search over a finite range \cite{zhang2025qronos, zhang2024magr}, or through simple heuristics, such as line search over the mean squared error between full-precision and quantized weights \cite{ashkboos2024quarot} or pre-activations \cite{gong2024llmc}.


{\bf Contributions.}
To our knowledge, no backpropagation-free algorithm currently performs per-channel quantization while automatically determining scaling factors. They all require a separate selection step for the scaling factors. In this work, we propose Beacon, an algorithm that carries out per-channel PTQ directly on the unscaled $b$-bit grid
\(
\mathcal{A}  = \{z+k : k=0,\dots,2^b\!-\!1\}
\)
and  determines the optimal scaling at the end by exploiting the geometry of scalar quantization. Like other state-of-the-art PTQ methods, Beacon requires only a small calibration set and avoids backpropagation.

\section{Preliminaries on Asymmetric Per-channel Quantization}

Given a calibration matrix $X\in\R^{m\times N}$, a PTQ algorithm typically seeks a quantized weight matrix $Q$ that minimizes the layer-wise reconstruction error. 
Specifically, in per-channel quantization, each column $w \in \mathbb{R}^N$ of $W$ is associated with its own scaling factor, so that each column of $Q$ is drawn from an integer grid $\mathcal{A}$ and rescaled by a diagonal matrix of scales $s \in \mathbb{R}^{N'}$. Thus the goal is to solve
\begin{equation}
    \min_{Q \in \mathcal{A}^{N \times N'}, \ s \in \mathbb{R}^{N'}} \ \|XW - XQ \mathrm{Diag}(s)\|_F^2.
\end{equation}
Here we slightly abuse our notation by using the same symbol $\mathcal{A}$ for all columns, when each column has its own zero point given by \eqref{eq:zero point}. Because the Frobenius norm decomposes as a sum of squared column errors, the problem separates across channels, which can be handled in parallel. Thus, it suffices to study a single column $w \in \mathbb{R}^N$, leading to the per-channel objective
\begin{equation}\label{eq:opt problem}
    \min_{q \in \mathcal{A}^N,\ c \in \mathbb{R}} \ \|Xw - cXq\|_2^2.
\end{equation}
Although this problem is NP-hard in general \cite{hassibi2002expected}, the subproblem of optimizing $c$ for a fixed $q$ admits a closed-form solution.




\begin{proposition}\label{thm:c given q}
    For any $q$, the optimal $c$ associated with the least squres objective in \eqref{eq:opt problem}  is 
    \begin{equation}\label{eq:c given q}
c = \frac{\langle Xw, Xq\rangle}{\|Xq\|^2}.
\end{equation}
    \begin{proof}
        \cref{eq:opt problem} is a least square problem in the one-dimensional real variable $c$ when $q$ is fixed. Taking derivatives with respect to $c$, the optimality condition is $2\|Xq\|^2 c - 2 \langle Xw, Xq\rangle = 0$, which implies the result.
    \end{proof}
\end{proposition}

\begin{corollary}\label{cor:opt sol prop}
    The global optimizer $(c^*, q^*)$ to \eqref{eq:opt problem} must satisfy the fixed point equation 
    \begin{equation}\label{eq:fixed pt eq}
c^* = \frac{\langle Xw, Xq^*\rangle}{\|Xq^*\|^2}.
\end{equation}
\end{corollary}

\section{Beacon: PTQ with Automatic Per-channel Scaling}

By \cref{thm:c given q}, the optimal scaling constant $c$ for a chosen $q$ is uniquely given by \eqref{eq:c given q}. Substituting this into \eqref{eq:opt problem} eliminates $c$ and yields the equivalent objective
\[
    \min_{q \in \mathcal{A}^N} \ \left\|Xw - \frac{Xq(Xq)^\top Xw}{\|Xq\|^2}\right\|^2.
\]
Since
\(
    Xw - \tfrac{Xq(Xq)^\top Xw}{\|Xq\|^2} = (I-\mathcal{P}_{Xq})(Xw),
\)
where $\mathcal{P}_{Xq}=\frac{1}{\|Xq\|^2}Xq(Xq)^\top$ is the projection onto $\mathrm{span}(Xq)$, the problem reduces to
\[
   \max_{q \in \mathcal{A}^N} \ \|\mathcal{P}_{Xq}(Xw)\|^2 
   \;=\; \max_{q \in \mathcal{A}^N} \ \frac{|\langle Xw, Xq\rangle|^2}{\|Xq\|^2}.
\]
%
%
%
Since $Xw$ is fixed, the objective is further equivalent to
\[
    \max_{q \in \mathcal{A}^N} \ |\cos\angle(Xw,Xq)|,
\]
where
\(
    \cos\angle(Xw,Xq) = \frac{\langle Xw, Xq\rangle}{\|Xw\|\cdot\|Xq\|}.
\)
We can drop the absolute value and attempt to solve
\[
    \max_{q \in \mathcal{A}^N} \ \cos\angle(Xw,Xq).
\]
This formulation reveals the geometry of scalar quantization:  aligning the directions of  $Xq$ and $Xw$ is \emph{all you need}.


{\bf Beacon. }Inspired by the greedy path-following algorithm of \citet{lybrand2021greedy}, Beacon starts by adopting a greedy approach to sequentially assign each $q_i \in \mathcal{A}$ for $i \in [N]$. Suppose $q_1^{(0)},\dots,q_{t-1}^{(0)}$ have already been chosen. The next coordinate is initialized by
\[
    q_t^{(0)} \in \arg\max_{p \in \mathcal{A}} \ \cos\angle\!\big(X_{\leq t} w_{\leq t}, \, X_{\leq t-1} q_{\leq t-1}^{(0)} + X_t p\big).
\]
This produces an initial vector $q^{(0)} \in \mathcal{A}^N$ with objective value
\[
    e_0 = \cos\angle(Xw, Xq^{(0)}).
\]

We then refine $q$ by cyclically updating each coordinate. At step $t$ of the $\ell$th cycle (loop), with all other coordinates fixed, the update rule is that $q_t^{(\ell)} $ must belong to
\[
 \arg\max_{p \in \mathcal{A}} \ \cos\angle\!\big(Xw, \, X_{\leq t-1} q_{\leq t-1}^{(\ell)} + X_t p + X_{\geq t+1} q_{\geq t+1}^{(\ell-1)}\big).
\]
 After the $\ell$-th full sweep of updates, we denote the resulting vector by $q^{(\ell)} \in \mathcal{A}^N$ and the corresponding objective value by
\[
    e_\ell = \cos\angle(Xw, Xq^{(\ell)}).
\]

\begin{proposition}
    The sequence $( e_\ell )_{\ell=0}^\infty$ converges in a finite number of iterations.
    \begin{proof}
        By design of the $q$ update procedure, $( e_\ell )_{\ell=0}^\infty$ is a non-decreasing sequence satisfying $0\leq e_\ell\leq 1$. By the monotone convergence theorem, it must converge. Moreover, since $q \in \mathcal{A}^N$, the sequence $( e_\ell )_{\ell=0}^\infty$ can take at most $|\A|^N$ distinct values. Let $\mathcal{I}=\{\ell\in\mathbb{N}: e_\ell>e_{\ell-1}\}$, then $\mathcal{I}$ must be a finite subset of natural numbers. Let $\ell^*$ be the maximal index for which $e_\ell>e_{\ell-1}$, and we have $e_L = e_{\ell^*}$ for all $L\geq \ell^*+1$.
    \end{proof}
\end{proposition}
Our experiments suggest that the quality of the quantized model improves during the first few $\ell$-loops and then plateaus, with the best results typically reached after 4--6 loops. The final step of Beacon is to use the resulting $q$ to compute the optimal scaling factor $c \in \mathbb{R}$ via \eqref{eq:c given q}.

We also empirically observe that sorting the columns of $X$ by increasing $\ell_2$ norm order in the initial assignment of $q^{(0)}$ (and by decreasing $\ell_2$ norm order in the refinement updates) improve the result upon natural index order. We present an intuition behind this observation. Let $u=\frac{X_{\leq t} w_{\leq t}}{\|X_{\leq t} w_{\leq t}\|}$ and $v=X_{\leq t-1} q_{\leq t-1}^{(0)}$. When determining $q_t^{(0)}$, we are selecting $p\in\A$ to maximize $\langle u, \frac{v+X_t p}{\|v+X_t p\|}\rangle$. If $\|X_t\|$ is dominated by $\|v\|$, $\frac{v+X_t p}{\|v+X_t p\|}$ will be largely determined by $\frac{v}{\|v\|}$, leaving little room for optimization. Thus, having the columns of $X$ in increasing $\ell_2$ norm order aligns better with  Beacon. 

{\bf Memory Efficient Implementation. }We observe that the angle between $Xw$ and $Xq$ is rotation invariant. Let $X=UR$ be the QR decomposition of $X$. Then we have $\cos\angle(Xw,Xq)=\dfrac{\langle Rw, Rq\rangle}{\|Rw\|\cdot\|Rq\|}$, which reduces the problem from dealing with an often very tall matrix $X\in \R^{m\times N}$ to a square matrix $R\in \R^{N\times N}$.

\textbf{Handling Error Accumulation. } Quantizing weights in earlier layers affects the inputs to subsequent layers. Let $X \in \mathbb{R}^{m \times N}$ denote the calibration set of $m$ samples (\textit{e.g.}, tokens) from the original pre-trained model, and let $\widetilde{X} \in \mathbb{R}^{m \times N}$ denote its counterpart from the partially quantized model. To account for the propagation of quantization error \cite{zhang2025qronos}, one can address the mismatch between $X$ and $\widetilde{X}$ by approximately solving
\[
    \min_{Q \in \mathcal{A}^{N \times N'}, \ s \in \mathbb{R}^{N'}} \ \|XW - \widetilde{X}Q \mathrm{Diag}(s)\|_F^2.
\]
Beacon can be generalized in a memory-efficient way to handle distinct inputs $X$ and $\widetilde{X}$. Let $\widetilde{X} = UR$ be the QR decomposition of $\widetilde{X}$. Then we seek

\begin{equation}
\begin{aligned}
q &\in \underset{p \in \mathcal{A}^N}{\argmax} \ \frac{\langle Xw, \widetilde{X}p\rangle}{\|Xw\|\cdot\|\widetilde{X}p\|} = \underset{p \in \mathcal{A}^N}{\argmax} \ \frac{\langle Xw, \widetilde{X}p\rangle}{\|\widetilde{X}p\|} \\
  &
  = \underset{p \in \mathcal{A}^N}{\argmax} \ \frac{\langle (U^\top X)w, Rp\rangle}{\|U^\top Xw\|\cdot\|Rp\|}.
\end{aligned}
\end{equation}


Thus, the problem again reduces from approximately solving  
\(
    \max_{q \in \mathcal{A}^N} \ \cos\angle(Xw,\widetilde{X}q)
\)
to 
\(
    \max_{q \in \mathcal{A}^N} \ \cos\angle(U^\top Xw,Rq),
\)
so that we work with the square matrices $U^\top X$ and $R$ rather than the potentially tall matrices $X$ and $\widetilde{X}$. We refer to this variant as \textbf{Beacon with error correction}. Its memory-efficient implementation is summarized in the algorithm below. In the special case without error correction, there is only one input $X=UR$, and we simply set $L=\widetilde{L}=R$.

\begin{algorithm}
\caption{\textbf{Beacon} w/ Error Correction.}
\label{main alg}
\begin{algorithmic}[1]
\State \textbf{Input:} $X, \widetilde{X}, W, b, \ell_{\max}$
\State $\widetilde{X}=UR$ \Comment{Compute QR decomposition}
\State $L=U^\top X$, $\widetilde{L}=R$
\State  $\A^0:=\{0,1,\dots,2^b\!-\!1\}$
\For{every $w$ in $W$ in parallel}
\State $\A =\A^0 + \left\lceil\frac{\min(w)}{\max(w)-\min(w)} \cdot (2^b-1)\right\rfloor$
\State  \( q^{(0)} = \mathbf{0}^{N} \)
\For{$t = 1$ to $N$} 
        \State 
        \(
        q_{t}^{(0)} \in \arg \max_{p\in\A} \cos\angle(L_{\leq t}w_{\leq t},\widetilde{L}_{\leq t-1}q_{\leq t-1}^{(0)}\newline \text{\qquad \qquad\qquad\qquad\qquad\qquad}+\widetilde{L}_t p)
        \)
    \EndFor
\For{ $\ell = 1$ to $\ell_{\max}$}
    \For{$t = 1$ to $N$} 
        \State $q_{t}^{(\ell)}\in \arg \max_{p\in\A} \cos\angle(Lw,\widetilde{L}_{\leq t-1}q_{\leq t-1}^{(\ell)}\newline \text{\qquad \qquad\qquad\qquad\qquad}+\widetilde{L}_t p+\widetilde{L}_{\geq t+1}q_{\geq t+1}^{(\ell-1)}))\}$
    \EndFor
\EndFor
\State $c = \frac{\langle Lw, \widetilde{L} q^{(\ell_{\max})}\rangle}{\|\widetilde{L}q^{(\ell_{\max})}\|^2}$
\State \Return $cq^{(\ell_{\max})}$
\EndFor
\end{algorithmic}
\end{algorithm}


{\bf Normalization Tuning. }A common practice in PTQ is to add a lightweight training step to tune the unquantized parameters in batch normalization (BN) or layer normalization (LN) layers, helping to compensate for quantization error. We evaluate the effect of normalization tuning in our experiments.

\section{Experiments}
We first test {\bf Beacon} on the DeiT-B vision transformer model (\cite{touvron2021training}, 86 million parameters) on the benchmark ImageNet classification task. To that end, we use DeiT-B ($patch16\_224$ version) from the Hugging Face timm library \cite{wightman2021resnet}. We focus on ILSVRC-2012 \cite{deng2009imagenet}, a 1000-category dataset with 1.28 million training images and 50 thousand validation images. All images in ILSVRC-2012 are preprocessed in the standard manner by resizing each image to $256\times256 $ and using the normalized $224\times224$ center crop. 
We evaluate top-1 accuracy of the quantized models on the entire validation set. The original accuracy of DeiT-B is $81.74\%$. We use batch size $2048$ to generate the calibration data. We ran our experiments on a single Nvidia A100 GPU with 80G GPU memory.

\begin{table}[ht!]
\centering
\caption{\textbf{Weight-only quantization of DeiT-B.}}
\begin{tabular}{cc}
\toprule
 & Beacon w/o E.C. (before/after LN)  \\
\midrule
1.58-bit(K=6) & 67.49 / \bf 72.82   \\
\midrule
2-bit(K=6) & 75.01 / \bf 77.19   \\
\midrule
3-bit(K=5) & 80.53 / \bf 80.69   \\
\midrule
4-bit(K=4) & 81.35 / \bf 81.40   \\
\bottomrule
runtime & $1-2\times$  \\
\bottomrule
\end{tabular}
\label{tab:deitb_quant}
\end{table}

\cref{tab:deitb_quant} displays the result of  quantizing DeiT-B via Beacon without error correction, as the number of bits varies. $1.58$-bit quantization means the grid is a scaling and shifting of $\{0, 1, 2\}$ per channel. We display results before and after a lightweight LN tuning after the whole model has been quantized. $K=\ell_{\max}$ is the number of loops we applied, and is fixed for each row. We remark that the best $K$, numerically, for each variant (whether with or without error correction or LN tuning, the bit width) slightly differ, but are typically  $K=4,5,$ or $6$. The runtime, compared to GPTQ with the same set up and machine, is reported in the last row. 
The LN tuning step only adds a small extra cost when training for $1$ epoch with $100\!-\!200$ data batches with batch size $128$.

We compare {\bf Beacon} to a recent state-of-the-art method for quantizing vision models, namely COMQ \cite{zhang2025comq}, and the standard baseline method GPTQ \cite{frantar2022gptq} in \cref{tab:compare comq}, again evaluated on DeiT-B. We implement GPTQ with asymmetric quantization on a standard per-channel min-max grid ($\alpha\!=\!\beta\!=\!1$, $z$ given by \eqref{eq:zero point}) and use the reported result in \citet{zhang2025comq}. As the original accuracy of DeiT-B reported in \citet{zhang2025comq} is $81.99\%$, which differs slightly from the $81.74\%$ we observe, here we compare the accuracy drop associated with each algorithm. The results suggest that fine grid tuning is critical to $<3$ bit quantization, further highlighting the benefit of {\bf Beacon} as a tuning free method for setting the scale. Indeed, {\bf Beacon} achieves the best performance for the challenging 2-bit case. We also note from the open-sourced code that \citet{zhang2025comq} used the entire training dataset for LN tuning and carefully initialize the scaling parameters to be $\alpha\!=\!\beta\!=\!0.65$ for 2-bit case. In contrast, Beacon only uses a few batches for LN tuning and, by construction, requires no hyper-parameter tuning. 

\begin{table}[ht!]
\centering
\caption{\textbf{Accuracy Drop(\%) Comparison on DeiT-B.}}
\begin{tabular}{cccc}
\toprule
 & 2-bit & 3-bit & 4-bit \\
\midrule
GPTQ (be/af LN) & 20.48 / 15.15 & 1.81 / 1.56 &  0.42 / {\bf 0.32} \\
\midrule
COMQ & 4.85 & 1.52 & 0.59 \\
\midrule
Beacon & \bf 4.55 & \bf 1.05 & \bf 0.34\\
\bottomrule
\end{tabular}
\label{tab:compare comq}
\end{table}

\cref{tab:deit3 large} displays the result of quantizing  DeiT-III-L (\cite{touvron2022deit}, 304 million parameters, original accuracy 84.59\%) with LN tuning for all methods. The first two columns are for {\bf Beacon} with $K=6$, with and without error correction respectively. GPTQ is still implemented with asymmetric quantization on a per-channel min-max grid. 
\begin{table}[ht!]
\centering
\caption{\textbf{Accuracy(\%) on DeiT-III-L.}}
\begin{tabular}{ccccc}
\toprule
 &  w/o E.C. &  w/ E.C. & GPTQ & GPTQ*\\
\midrule
2-bit & 80.67 & \bf 80.76 & 78.10 & 78.28\\
\midrule
1.58-bit & 77.07 & \bf 77.32 & 46.25 & 72.74\\
\bottomrule
\end{tabular}
\label{tab:deit3 large}
\end{table}
GPTQ* uses {\bf Beacon} to generate its scaling factors: Given a channel $w$ and its unscaled grid $\mathcal{A} = \{z\!+\!k : k=0,\dots,2^b\!-\!1\}$ with $z=\left\lceil\frac{\min(w)}{\max(w)-\min(w)} \cdot (2^b-1)\right\rfloor$, we first run {\bf Beacon} without error correction, giving us a scaling factor $\hat{c}$ for this channel to replace the choice \(
c = \frac{\max(w) - \min(w)}{2^b-1}
\) used in GPTQ. We observe that the scaling generated by {\bf Beacon} significantly improves the 1.58-bit GPTQ quantized model to a usable quality, although {\bf Beacon} still outperforms it. \cref{tab:deitb_quant} and \cref{tab:deit3 large} both demonstrate that {\bf Beacon} yields a usable model even in the extremely challenging 1.58-bit quantization setting.

\section{Conclusion}
We introduced {\bf Beacon}, a simple and tuning-free algorithm for post-training quantization. Unlike existing approaches that require heuristic scale selection or iterative search, Beacon directly quantizes weights using an unscaled grid and infers optimal scaling factors per-channel after quantization. This streamlines the PTQ process and avoids dependence on hyperparameter tuning or back-propagation. Beacon yields models that retain compatibility with standard hardware and matches the performance of more complex state-of-the-art methods, despite its minimal calibration and computational overhead. We believe that this makes it a simple and effective solution for compressing large models in resource-constrained environments.


\bibliographystyle{abbrvnat}
\bibliography{citations}

\end{document}